\documentclass[10pt,twocolumn,letterpaper]{article}

\usepackage{cvpr}              

\usepackage{algorithm}
\usepackage{algorithmic}

\newcommand{\subdivision}[1]{%
  \vspace{3pt}
  \noindent\textbf{#1}
  \noindent
}

\newcommand{\fig}{{Fig.}}
\newcommand{\tab}{{Table}}

\usepackage{bm}
\usepackage{tikz}
\usepackage{amsmath} 
\usepackage{amssymb}
\usepackage{array}
\usepackage{booktabs} 
\usepackage{tabularx} 
\usepackage{multirow} 
\usepackage{hhline}
\usepackage{tcolorbox}
\usepackage{cellspace}
\usepackage{colortbl}
\usepackage{xcolor}
\definecolor{highlight}{HTML}{FFC000}

\usepackage[accsupp]{axessibility}  

\usepackage{siunitx}
\usepackage{caption}
\usepackage{wrapfig}
\usepackage{adjustbox}
\usepackage{ragged2e} 
\usepackage{hyphenat}      
\usepackage[numbers,sort&compress]{natbib}

\definecolor{cvprblue}{rgb}{0.21,0.49,0.74}
\usepackage[pagebackref,breaklinks,colorlinks,allcolors=cvprblue]{hyperref}

\def\paperID{398} 
\def\confName{CVPR}
\def\confYear{2025}

\title{Dynamic Integration of Task-Specific Adapters for Class Incremental Learning}

\author{
    Jiashuo Li\textsuperscript{\rm 1},
    Shaokun Wang\textsuperscript{\rm 1}\textsuperscript{\rm \( \dagger \)},
    Bo Qian\textsuperscript{\rm 1},
    Yuhang He\textsuperscript{\rm 2},
    Xing Wei\textsuperscript{\rm 1},
    Qiang Wang\textsuperscript{\rm 1},
    Yihong Gong\textsuperscript{\rm 1,2}\textsuperscript{\rm \( \dagger \)} \\
    \textsuperscript{\rm 1} School of Software Engineering, Xi'an Jiaotong University \\
    \textsuperscript{\rm 2} College of Artificial Intelligence, Xi’an Jiaotong University \\
    {\tt\small \{xjtuljs,qb990531,qwang\}@stu.xjtu.edu.cn; shaokunwang.xjtu@gmail.com;} \\ 
    {\tt\small heyuhang@xjtu.edu.cn; \{weixing,ygong\}@mail.xjtu.edu.cn}
}

\begin{document}
\maketitle
\begin{abstract}
Non-exemplar Class Incremental Learning (NECIL) enables models to continuously acquire new classes without retraining from scratch and storing old task exemplars, addressing privacy and storage issues.
However, the absence of data from earlier tasks exacerbates the challenge of catastrophic forgetting in NECIL. 
In this paper, we propose a novel framework called Dynamic Integration of task-specific Adapters (DIA), which comprises two key components: Task-Specific Adapter Integration (TSAI) and Patch-Level Model Alignment.
TSAI boosts compositionality through a patch-level adapter integration strategy, aggregating richer task-specific information while maintaining low computation costs. 
Patch-Level Model Alignment maintains feature consistency and accurate decision boundaries via two specialized mechanisms: Patch-Level Distillation Loss (PDL) and Patch-Level Feature Reconstruction (PFR). 
Specifically, on the one hand, the PDL preserves feature-level consistency between successive models by implementing a distillation loss based on the contributions of patch tokens to new class learning. 
On the other hand, the PFR promotes classifier alignment by reconstructing old class features from previous tasks that adapt to new task knowledge, thereby preserving well-calibrated decision boundaries. 
Comprehensive experiments validate the effectiveness of our DIA, revealing significant improvements on NECIL benchmark datasets while maintaining an optimal balance between computational complexity and accuracy. 
\end{abstract}    
\let\thefootnote\relax\footnote{\( ^\dagger \) Yihong Gong and Shaokun Wang are the corresponding authors.}%
\section{Introduction}
\label{sec:intro}

\begin{figure}[t!]
    \centering
    \includegraphics[width=0.9\linewidth]{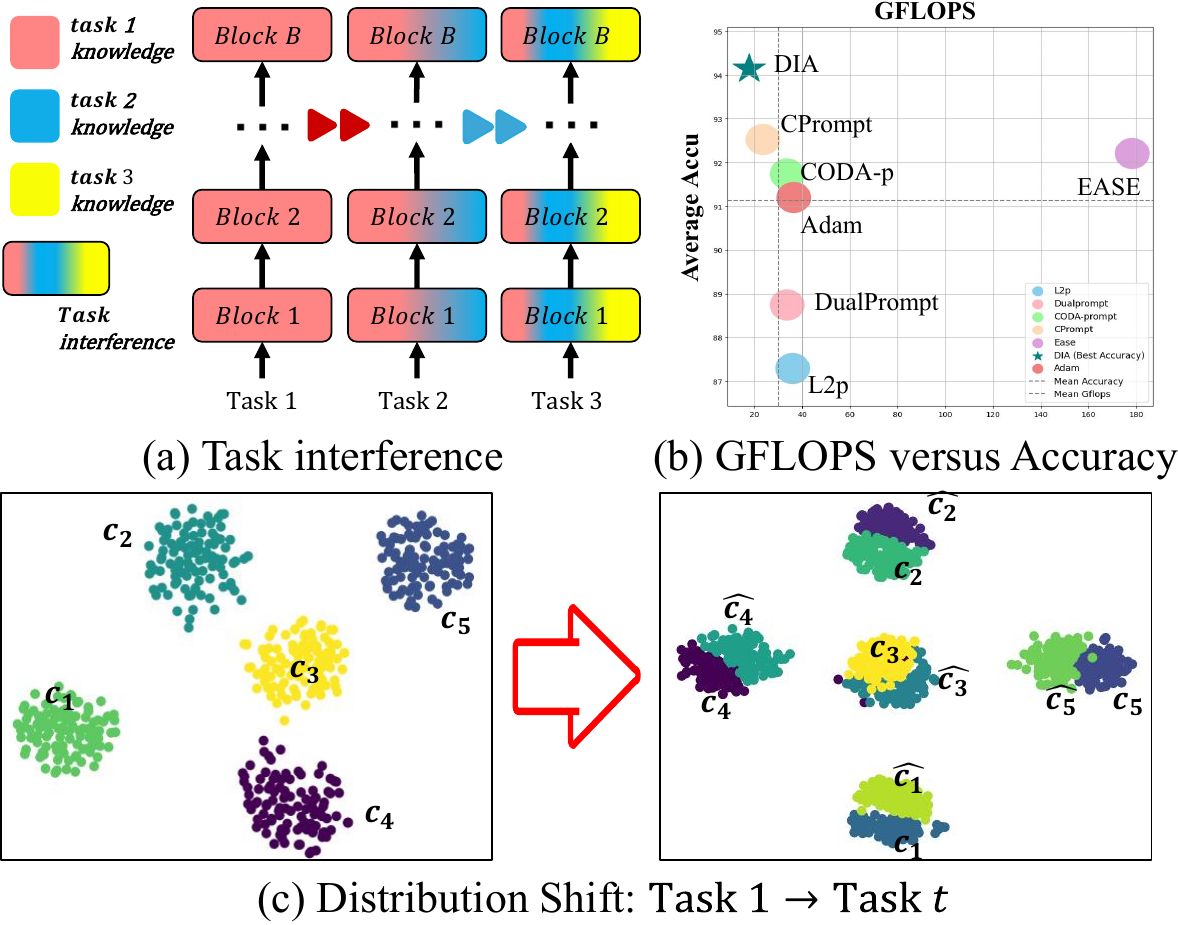}
    \caption{(a) Shared parameter space leads to task interference. (b) High computation costs characterize current PET-based methods. (c) 
    Gaussian-based feature reconstruction may have a significant deviation from the actual feature distribution. \( c_i \) indicates the actual feature distribution of old classes, \( \hat{c_i} \) indicates the feature distribution generated by Gaussian Sampling.}
    \label{fig: head}
\end{figure}

Class Incremental Learning (CIL) has gained considerable attention within the expansive field of artificial intelligence research~\cite{qiao2024class,2021CVPRPrototype_Zhu,2022CVPRLearning_Wang,kurniawan2024evolving,li2024towards,ijcai2024p346,OSP,ICCVW,2023TCSVTSemantic_Wang,bonato2024mind,zhai2024AAAIFineGrained2024}, offering the potential for models to continuously learn new knowledge while maintaining the knowledge of previously encountered old classes.
Replay-based CIL approaches~\cite{2021CVPRPrototype_Zhu,Learningforgetting_Li,FOSTERFeatureBoosting_Wang,model603exemplars_Zhou,iCaRLIncrementalClassifier_Rebuffi,dynamicallyexpandablerepresentation_Yan} require the storage of exemplars from previous tasks, which presents challenges in terms of memory limitations and privacy concerns. 
Benefiting from the advances in Pre-Trained Models (PTMs)~\cite{imageworth16x16_Dosovitskiy}, Non-Exemplar Class Incremental Learning (NECIL) has emerged as a promising alternative, enabling the incremental acquisition of knowledge without the need to maintain a buffer of old class exemplars. 

Despite the inherent ability of PTMs to produce generalizable features, which has led to superior performance in the NECIL setting, NECIL remains challenging due to two primary factors. 
\textbf{First}, as shown in \fig\ref{fig: head} (a), studies~\cite{2023ICCVunified_Gao, 2023ICCVSLCA_Zhang, 2024arXivContinual_Zhou} that tune a shared parameter space for all tasks are unable to segregate parameters for each task. 
The mixing of parameters leads to task interference, which in turn causes catastrophic forgetting. 
Particular PET-based methods~\cite{2022ECCVDualPrompt_Wang,2022CVPRLearning_Wang,2023CVPRCODAprompt_Smitha,2024CVPRConsistent_Gao, kurniawan2024evolving} isolate task parameters by designing task-specific prompts.
However, these methods establish a strong correlation between prompt selection and the class tokens. 
This tight coupling neglects the rich semantic information embedded in patch tokens and the task-specific knowledge preserved in the intact old task parameters, thereby undermining the model's capacity for effective knowledge retention and reproduction in NECIL scenarios.
Moreover, these methods, requiring multiple forward propagations, increase computational costs, as shown in \fig\ref{fig: head} (b). 
\textbf{We summarize the above issues as the compositionality deficiency.} 

{\textbf{Second}, the absence of exemplars from previous tasks prevents the model from utilizing replay techniques, thereby posing significant challenges in maintaining feature consistency and preserving accurate decision boundaries.} 
Current PET-based methods~\cite{2022ECCVDualPrompt_Wang,2022CVPRLearning_Wang,2023CVPRCODAprompt_Smitha,2024CVPRConsistent_Gao} overlook the necessity of maintaining consistency across incrementally trained models. 
{In incremental learning scenarios, the representations of old task samples will change due to the incorporation of new task knowledge, and this neglect ultimately leads to feature drift in the old task samples.}
Additionally, these approaches are unable to adapt classifiers to the decision boundary changes introduced by new tasks. 
{Regularization-based approaches}~\cite{2024AAAIFineGrained_zhai,2023ACMMPolo_wang,2021CVPRPrototype_Zhu,2022SSRE,2018LwF} that impose constraints on class tokens \textit{to maintain feature consistency} often result in overly restrictive regularization, hindering the learning for new tasks.
{Prototype-based approaches}~\cite{2023ICCVSLCA_Zhang,2021CVPRPrototype_Zhu,2023ACMMPolo_wang} leverage Gaussian distributions to generate previous task features and then align classifiers. However, it can \textit{lead to inaccurate decision boundaries}, as features generated by Gaussian sampling progressively diverge from the actual features as new tasks are introduced, as shown in \fig\ref{fig: head} (c). 
\textbf{We attribute the above issues to the model alignment deficiency.} 

To mitigate the above two challenges, we introduce the Dynamic Integration of task-specific Adapters (DIA) framework, which comprises two components: Task-Specific Adapter Integration (TSAI) and Patch-Level Model Alignment. 
Specifically, the TSAI module is designed to boost compositionality through a patch-level adapter integration strategy, which provides a more flexible compositional solution while maintaining low computation costs. 
Moreover, we demonstrate the knowledge retention and reconstruction potential of TSAI on a parameter factorization basis. 
Building upon these capabilities, we propose a patch-level model alignment strategy to preserve feature consistency and maintain accurate decision boundaries across tasks.
This strategy integrates two fundamental parts:
1) The \textbf{Patch-Level Distillation Loss (PDL)} ensures feature-level consistency between the old and new models by introducing a distillation loss that regulates the feature shift of patch tokens. PDL evaluates the contribution of patch tokens to the new task learning and then penalizes feature drift in non-contributory tokens to maintain feature consistency.
2) The \textbf{Patch-Level Feature Reconstruction (PFR)} retrieves patch tokens that encapsulate knowledge from prior tasks and combines them with prototypes to reconstruct old class features aligned with the newly learned tasks. 
These reconstructed features facilitate classifier alignment, ensuring the maintenance of an accurate decision boundary throughout the incremental process.

Comprehensive experiments on four NECIL benchmark datasets demonstrate that our DIA method achieves state-of-the-art (SOTA) performance. 
Moreover, as shown in \fig\ref{fig: head} (b), our DIA maintains up to a 90\% decrease in computational complexity while maintaining SOTA performance. In summary, the main contributions include: 
\begin{itemize}
    \item We propose a novel framework entitled Dynamic Integration of task-specific Adapters (DIA) for the NECIL problem, addressing compositionality and model alignment deficiencies.
    \item We introduce the Task-Specific Adapter Integration (TSAI) module to boost compositionality, which employs a patch-level adapter integration strategy. We also demonstrate its knowledge retention and reconstruction capability through parameter factorization analysis.
    \item We present a patch-level model alignment strategy based on TSAI:
    {PDL maintains feature-level consistency by regulating feature drift in non-contributory patch tokens. PFR improves classifier alignment by reconstructing old class features aligned with newly learned tasks.}
    \item Experimental results across four benchmarks demonstrate that our DIA achieves SOTA performance while maintaining an optimal balance between computational complexity and accuracy.
\end{itemize}

\section{Related Work}

\subdivision{Parameter-Efficient Tuning:} 
As an efficient alternative to full fine-tuning, Adapter Tuning~\cite{2019ICMLParameterefficient_houlsby} was initially introduced to efficiently transfer large pre-trained models to downstream tasks in NLP tasks. Afterward, methods such as Prompt-Tuning~\cite{2021EMNLPPower_lester,2021IJCNLPPrefixtuning_li} adapt models by inserting learnable tokens explicitly tailored to the new tasks. Following the success of Vision Transformers~\cite{imageworth16x16_Dosovitskiy,2021ICCVSwin_liu}, PET methods have been adapted for visual transfer learning. Notable examples include Visual Prompt Tuning (VPT)~\cite{2022ECCVVisual_jia} and AdapterFormer~\cite{2022NIPSAdaptFormer_Chen}, which apply PET techniques to vision tasks. These methods achieve comparable or superior performance to full fine-tuning while maintaining efficiency. 
In this paper, we propose a patch-level task adapter integration strategy that enables patch tokens to aggregate richer task information with greater flexibility while maintaining low computational costs.

\begin{figure*}[t!]
    \centering
    \includegraphics[width=0.9\linewidth]{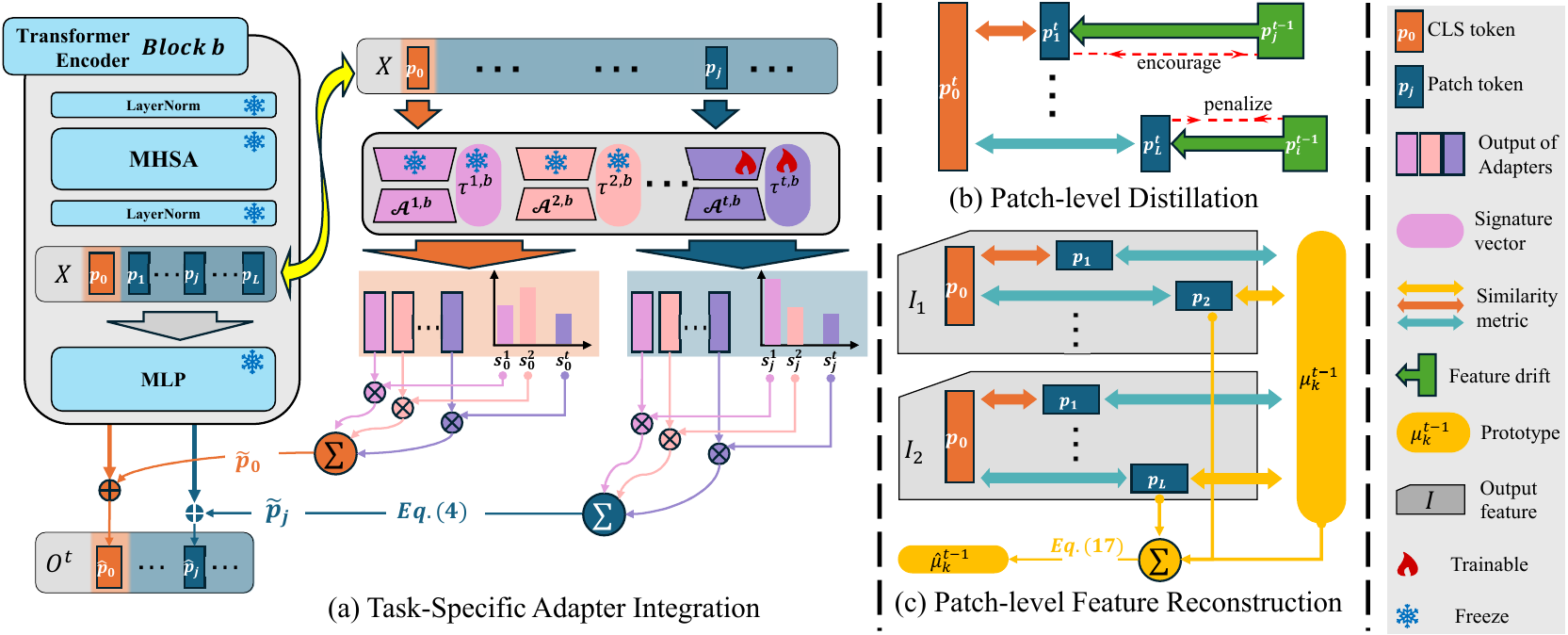}
    \caption{Illustration of DIA. (a) \textbf{Task-Specific Adapter Integration}. For incremental task \( t \), we learn a task adapter \( \mathcal{A}^{t,b} \) and a task signature vector \( \bm{\tau}^{t,b} \in \mathcal{R}^d \) at each transformer block \(b\).
    Each token is routed to the relevant task adapters through the signature vectors, processed independently, and combined into an integrated, task-informed output.
    (b) \textbf{Patch-level Distillation}. We promote feature drift in patch tokens that contribute to new task learning while penalizing those that do not, thus regulating the feature shift associated with old tasks. (c) \textbf{Patch-level Feature Reconstruction}. We identify \textcolor{highlight}{patch tokens} that are related to old class knowledge and integrate them with the old class prototype \( \bm{\mu}_{k}^{t-1} \) to reconstruct old class feature \( \hat{\bm{\mu}}_{k}^{t-1} \) aligned with new tasks.}
    \label{fig: overview}
\end{figure*}

\subdivision{Non-exemplar Class Incremental Learning}
NECIL approaches address privacy and memory concerns by eliminating the need for old task exemplars and employing various techniques such as regularization~\cite{2018TPAMILearning_li,huang2024etag,2020CVPRSemantic_yu}, augmentation~\cite{2021NIPSClassincremental_zhu,2021CVPRPrototype_Zhu,kim2024cross}, and model-alignment strategies~\cite{DomainDrift,LDC2024,SDC2020,2023ACMMPolo_wang,2021CVPRPrototype_Zhu} to maintain model performance across tasks. 
Particularly, SDC~\cite{SDC2020} calculates class prototype shifts using feature gaps. LDC~\cite{LDC2024} uses a forward projector to learn mappings from old to new feature spaces.
Currently, PTM-based methods that sequentially adjust the PTM to stream data with new classes are a promising approach for NECIL. Most methods~\cite{2022CVPRLearning_Wang,2022ECCVDualPrompt_Wang,2023CVPRCODAprompt_Smitha,2024CVPRConsistent_Gao,2024CVPRConvolutional_Roy} focus on the instance-level ([cls] token-based) prompt selection to segregate task parameters. 
LAE~\cite{2023ICCVunified_Gao} and ADAM~\cite{2024arXivContinual_Zhou} propose unified frameworks for PET methods by model ensemble. 
C-ADA~\cite{PromptLearningContinual_gao} leverages Continual Adapter Layers and a Scale and Shift Module to learn new tasks.
SLCA~\cite{2023ICCVSLCA_Zhang} extends the Gaussian modeling~\cite{2021CVPRPrototype_Zhu} of old task features to rectify classifiers. 
EASE~\cite{2024CVPRExpandable_Zhou} utilizes adapters to extract task-specific features through multiple forward propagations. 
In this paper, the proposed DIA framework leverages TSAI to enhance compositionality and incorporates patch-level model alignment to mitigate feature shifts and decision boundary distortions.
\section{Problem Setup} 
NECIL involves sequentially learning a set of $T$ tasks, denoted as $\{\mathcal{T}^t\}_{t=1}^T$, where each task $\mathcal{T}^t = \{D^t, C^t\}$ consists of a current training set $D^t = \{(I^t_i, y^t_i)\}_{i=1}^{N^t}$ and a class label set $C^t = \{c^t_k\}_{k=1}^{M^t}$. In this context, $I^t_i$ is the input image, $y^t_i$ is the class label for the $i$-th image belonging to $C^t$, $N^t$ represents the number of images, and $M^t$ denotes the number of classes in $C^t$. There is no overlap between the classes of different tasks, meaning $\forall i,j,\ C^{i} \cap C^{j}=\varnothing$. 
After training on a task $\mathcal{T}^t$, the model is evaluated on all the classes encountered so far, $C^{1:t} = C^{1} \cup C^{2} \cup \cdots \cup C^{t}$. Notably, NECIL prohibits storing exemplars from old tasks. 

\section{Methodology}

\subsection{Overview}
\fig\ref{fig: overview} shows an overview of our DIA framework. 
At incremental task \( t \), we introduce a task-specific adapter \( \mathcal{A}^{t,b} \) in parallel with the MLP layer, along with a task signature vector \( \bm{\tau}^{t,b}\) into each transformer block \( b \). 
As shown in \fig\ref{fig: overview} (a), each image token is routed by the signature vectors \( [\bm{\tau}^{i,b}]_{i=1}^t \) to relevant adapters. 
These task-specific adapters independently process the input token, and their outputs are merged using scalars determined by the task signature vectors, yielding an integrated output. 
At the task \( t \), only \(\mathcal{A}^{t,b}\) and \(\bm{\tau}^{t,b}\) are trainable. 
Leveraging TSAI's capacity for knowledge retention and reconstruction, along with the rich information embedded in patch tokens,
we introduce a patch-level model alignment mechanism at both the feature and classifier levels. 
First, as shown in \fig\ref{fig: overview} (b), we compute the Patch-Level Distillation Loss (PDL) based on the feature drift between the patch tokens \(\mathbf{p}^{t}\) and \(\mathbf{p}^{t-1}\), which are obtained from the model trained on task \( t \) and task \( t-1 \), respectively. 
PDL penalizes the feature drift in patch tokens that do not contribute to the new task learning, thereby preserving the old task knowledge. 
Second, we propose a Patch-level Feature Reconstruction (PFR) method to reconstruct old class features without exemplars, as shown in \fig\ref{fig: overview} (c). 
We identify patch tokens related to old task knowledge and integrate them with the old task prototypes. The reconstructed features are then used to recalibrate the decision boundaries of the classifier, adapting to the changes in feature distribution.

\subsection{Task-Specific Adapters Integration}
Our TSAI aims to provide a more flexible compositional solution by employing a patch-level adapter integration strategy.
To achieve this goal, we learn a task-specific adapter \( \mathcal{A}^{t, b} \) and a task signature vector \( \bm{\tau}^{t, b} \in R^d \) for each incremental task, where \( d \) indicates feature dimension, \( t \) represents the incremental task id, and \( b \) represents the block index. For convenience, we omit block index \( b \), as the computation is identical across transformer blocks. 

Specifically, the task-specific adapter \( \mathcal{A}^{t} \) is a bottleneck module comprising a down-projection layer \( \mathbf{W}_{\text{down}} \in \mathbb{R}^{d \times r} \), an up-projection layer \( \mathbf{W}_{\text{up}} \in \mathbb{R}^{r \times d} \). 
This bottleneck module extends and adjusts the feature space by modifying the MLP output through the residual connection, utilizing a scaling vector \( \mathbf{s}^t \) derived from \( \bm{\tau}^{t} \). 
{For an input \( \mathbf{X} = {\left[{\mathbf{p}_j}^\top\right]}_{j=0}^{L} \in R^{(L+1) \times d} \), consisting of \( L+1 \) image tokens, where \( \mathbf{p}_0 \in \mathbb{R}^{d} \) indicates the class token and \( \mathbf{p}_{j} \in \mathbb{R}^{d}, j>0 \) denote the patch tokens.}
The task signature vector \( \bm{\tau}^{t} \) assigns scalar \( s_j^t \) to each image token \( \mathbf{p}_j \) to evaluate its task relevance:
\begin{gather}
\mathbf{\tilde{p}}_j^t = s_j^t \mathcal{A}^t(\mathbf{p}_j) = s_j^t \text{ReLU}(\mathbf{p}_j \mathbf{W}_{\text{down}}) \mathbf{W}_{\text{up}}, \\
s_j^t = < \bar{\mathbf{p}}_j,\bar{\bm{\tau}}^t >, \mathbf{s}^t = \left[ s_j^t \right]_{j=0}^L,
\end{gather}
where \(\bar{\mathbf{p}}_j, \bar{\bm{\tau}}^t\) represent normalized tensors \( \bar{\mathbf{p}}= \mathbf{p}/\|\mathbf{p}\|_2\), \( \bar{\bm{\tau}}^t= \bm{\tau}^t/\|\bm{\tau}^t\|_2 \), respectively, 
 and \( < \cdot, \cdot > \) represents the dot product. 

Subsequently, the output feature \(\mathbf{\hat{p}}_j\) for the token \(\mathbf{p}_j\) using the adapter \(\mathcal{A}^t \) can be calculated as follows:
\begin{align}
\mathbf{\hat{p}}_j &= \text{MLP}(\mathbf{p}_j) + \mathbf{\tilde{p}}_j^t + \mathbf{\mathbf{p}_j} \nonumber \\
                      &= \text{MLP}(\mathbf{p}_j) + s_j^t \mathcal{A}^t(\mathbf{p}_j) + \mathbf{p}_j.
\end{align}

For incremental task \( t > 1 \), the output feature \(\mathbf{\hat{p}}_j\) of TSAI is integrated from multiple task-specific adapters \( \{\mathcal{A}^i\}_{i=1}^t \). We apply the softmax operation to normalize the scaling vector \( [s_j^i]_{i=1}^t \) for the patch token \( \mathbf{p}_j \), enabling the model to combine the contributions from different adapters while maintaining a balanced and stable output.
\begin{gather}
\mathbf{\tilde{p}}_j^t = \sum_{i=1}^{t}{\hat{s}_j^i \mathcal{A}^i(\mathbf{p}_j)}, \\
\mathbf{\hat{p}}_j = \text{MLP}(\mathbf{p}_j) + \mathbf{\tilde{p}}_j^t + \mathbf{p}_j, \\
\mathbf{O}^{t} = {\left[ \mathbf{\hat{p}}_j^\top \right]}_{j=0}^{L}, \quad[\hat{s}_j^i]_{i=1}^t = \text{softmax}([s_j^i]_{i=1}^t),
\end{gather}
where \( \mathbf{O}^{t} \) is the output of TSAI for input \( \mathbf{X} \).

\subdivision{Knowledge Retention and Reconstruction Analysis:}
We conduct an in-depth analysis of the parameter space for each task-specific adapter, using matrix factorization to demonstrate that our TSAI module can effectively preserve and reproduce knowledge from previous tasks without needing exemplars. 
In particular, we start with no activation function to describe the knowledge retention mechanism under linear conditions and then extend to nonlinear conditions. 
{We provide a detailed analysis in the supplementary.}

To avoid confusion, we define the input token as \( \mathbf{p} \in \mathbb{R}^{d} \) and the adapter weight as \( \mathbf{W} = \mathbf{W}_{\text{down}}\mathbf{W}_{\text{up}} \in \mathbb{R}^{d \times d^{'}} \), with a matrix rank of \( r \). The weight matrix can be decomposed through SVD~\cite{1980SVD} without information loss:
\begin{align}
    \mathbf{W} &= \mathbf{U} \text{diag}(\sigma) \mathbf{V}, \quad \mathbf{W} = \sum \mathbf{u}_i \sigma_i \mathbf{v}_i^\top, \\
    \mathbf{U}^\top &= \left[ {\mathbf{u}_i}^\top \right]_{i=1}^r \in \mathbb{R}^{r \times d},
    \mathbf{V} = {\left[ {\mathbf{v}_i}^\top \right]_{i=1}^r} \in \mathbb{R}^{r \times d^{'}},
\end{align}
where \( \text{diag}(\sigma) \) is a diagonal matrix with singular values \(\sigma_i\) on the diagonal. \(\mathbf{U}\) is an \(d \times r\) orthogonal matrix, with the singular vectors \(\mathbf{u}_i\) as its columns. \(\mathbf{V}\) is an \( r \times d^{'} \) orthogonal matrix, with the singular vectors \(\mathbf{v}_i^\top\) as its rows. 

The output \( \mathbf{o} \) can be formulated as:
\begin{align}
 \mathbf{o} &= \mathcal{A}(\mathbf{p}) = \mathbf{W}^\top \mathbf{p} = \sum (\mathbf{u}_i \sigma_i \mathbf{v}_i^\top)^\top \mathbf{p} \nonumber \\
   &= \sum \mathbf{v}_i (\sigma_i \mathbf{u}_i^\top \mathbf{p})  = \sum \rho_{i} \mathbf{v}_i = \mathbf{V}^\top g(\mathbf{p}), \\
g&(\mathbf{p}) = {\text{diag}(\sigma)}^\top \mathbf{U}^\top \mathbf{p}= {[\rho_{i}]_{i=1}^r} \in \mathbb{R}^r.
\end{align}
\vspace{-\baselineskip}

From the above derivation, it is evident that each task adapter enables the model to learn task-specific ``keys'' and ``values'' independently of the input features. 
The output \( \mathbf{o} \) for each input token \( \mathbf{p} \) is obtained by weighting the task parameter space basis vectors \( \mathbf{v}_i \) through a linear function \(g\left(\cdot; \mathbf{U}, \text{diag}(\sigma)\right)\), which becomes non-linear in the presence of a ReLU activation. 
TSAI guarantees that the output discriminative information \( \mathbf{o} \) remains within the task subspace, irrespective of the input. This property can be effectively utilized by designing appropriate loss functions to maintain feature consistency under the NECIL setting.

\subsection{Patch-Level Model Alignment}
Due to the limitations of not storing old class exemplars, mainstream NECIL methods either do not perform model alignment or only align the classifier~\cite{2024CVPRExpandable_Zhou,2023ICCVSLCA_Zhang,2024CVPRConsistent_Gao}. 
Based on TSAI's knowledge retention and reconstruction capabilities, as well as the rich information conveyed by patch tokens, 
we propose a patch-level distillation loss to maintain feature consistency and a patch-level old class feature reconstruction method to align the classifier. 

\subdivision{Patch-Level Distillation Loss}:
To ensure that new tasks can share and reuse old task knowledge while maintaining consistent feature representations for old tasks, we propose patch-level distillation loss (PDL).

As shown in \fig\ref{fig: overview} (b), during the training of task \( t \), we obtain the output features \( \mathbf{X}^n \) and \( \mathbf{X}^o \)
from both the current model and the model trained on the previous task \(t-1\), each consists of \( L + 1 \) image tokens. 
\begin{align}
    \mathbf{X}^o = {\left[{\mathbf{p}_{j}^{t-1}}^\top\right]}_{j=0}^{L},
    \mathbf{X}^n = {\left[{\mathbf{p}_{j}^{t}}^\top\right]}_{j=0}^{L} \in \mathbb{R}^{(L+1) \times d},
\end{align}

We first compute the contribution of each patch token to the learning of the new task based on the angular similarity:
\begin{align}
    \alpha_{\cos} = \frac{\mathbf{p}_{0}^{t} \cdot \mathbf{p}_{j}^{t}}{\|\mathbf{p}_{0}^{t}\|_2 \|\mathbf{p}_{j}^{t}\|_2},
    \alpha_{\angle} = \frac{\pi - \arccos(-\alpha_{\cos})}{\pi},
\end{align}
where \( \alpha_{\cos} \) and \( \alpha_{\angle} \) represent the cosine similarity and angular similarity, respectively. We also conduct ablation experiments on different similarity metrics in \tab~\ref{Table: Metric}.

Afterward, we map patch tokens onto a hypersphere using \( L_2 \) normalization to ensure numerical stability. We measure the feature drifts \( \mathcal{D} \) by calculating the distances between the corresponding tokens on the hypersphere, as follows:
\begin{align}
    \mathcal{D}(\mathbf{p}_{j}^{t}, \mathbf{p}_{j}^{t-1}) = \| \bar{\mathbf{p}}_{j}^{t} - \bar{\mathbf{p}}_{j}^{t-1} \|_2.
\end{align}

We allow greater flexibility for patch tokens that contribute significantly to new tasks. For patches that contribute less to new tasks, we align their tokens with the output of the old model to maintain feature consistency with previous tasks. Finally, the PDL loss is defined as:
\begin{align}
    \mathcal{L}_{pdl} = \frac{1}{L} \sum_{j=1}^{L} \alpha_{\angle}(\mathbf{p}_{0}^{t}, \mathbf{p}_{j}^{t}) \mathcal{D}(\mathbf{p}_{j}^{t}, \mathbf{p}_{j}^{t-1}).
\label{eq: PDL}
\end{align}

\subdivision{Patch-Level Feature Reconstruction}:
As new incremental tasks are learned, the decision boundaries established from old tasks may undergo significant changes~\cite{2021CVPRPrototype_Zhu,2023ACMMPolo_wang}. 
Towards this end, we propose a patch-level feature reconstruction method to generate old class features that adapt to the evolving model, providing better classifier alignment.

Specifically, at each incremental task \( t - 1 \), we compute and store a class prototype \( \bm{\mu}_{k}^{t-1} \) for each class \( k \). When learning new incremental tasks, we apply a relative similarity difference  \( \delta_{(k, i, j)} \) to measure the relevance between patch token \( p_{(i, j)} \) and prototype \( \bm{\mu}_{k}^{t-1} \),  where \( p_{(i, j)} \) indicates the \(j\)-th token of \( X_i \) within the training batch. 
\begin{gather}
\delta_{(k,i,j)} = \frac{\alpha_{\text{cos}} (\bm{\mu}_{k}^{t-1}, \mathbf{p}_{(i,j)}) - \alpha_{\text{cos}} (\mathbf{p}_{(i,0)}, \mathbf{p}_{(i,j)})}{\alpha_{\text{cos}} (\bm{\mu}_{k}^{t-1}, \mathbf{p}_{(i,j)}) +\alpha_{\text{cos}} (\mathbf{p}_{(i,0)}, \mathbf{p}_{(i,j)})}, \\
\hat{\delta}_{(k,i, j)} = 
\begin{cases}
0, & \text{if } \delta_{(k,i, j)} \leq 0 \\
\delta_{(k,i, j)}, & \text{if } \delta_{(k,i, j)} > 0
\end{cases},
\end{gather}
{where \( \alpha_{cos} \) represents the cosine similarity.}

Afterward, we retrieve the patch tokens \(\mathbf{p}_{(i, j)}\) whose \( \hat{\delta}_{(k, i, j)} > 0 \) and integrate them with prototype \( \bm{\mu}_{k}^{t-1} \) using the convex combination to generate old task feature \( \hat{\bm{\mu}}_{k}^{t-1} \) as follows:
\begin{gather}
\hat{\bm{\mu}}_{k}^{t-1} = \beta \cdot \bm{\mu}_{k}^{t-1} + (1-\beta) \sum_{i, j} \omega_{(i, j)} \cdot \mathbf{p}_{(i, j)}, \\
\left[\omega_{(i,j)}\right]_{i,j} = \text{softmax}(\left[\hat{\delta}_{(k,i, j)}\right]_{i,j}),
\label{eq: PFR}
\end{gather}
where \( \beta \) is a hyperparameter.

In each training batch, we randomly generate old task features through our PFR and align the classifier through the cross-entropy loss \( \mathcal{L}_{CE} \).

\subsection{Optimization Objective}
We structure the optimization process into two distinct stages: \textbf{new task learning} and \textbf{classifier alignment}. To ensure feature consistency, the objective function for \textbf{new task learning} is defined as follows:
\begin{align}
    \mathcal{L}_{obj} = \mathcal{L}_{CE} + \lambda \mathcal{L}_{pdl},
\end{align}
where \( \lambda \) is a hyperparameter to balance the contributions of the two losses. 

To further refine the classifier, \textbf{classifier alignment} is executed after new task learning, following the pipeline detailed in~\cite{2023ICCVSLCA_Zhang,2021CVPRPrototype_Zhu}. Specifically, during the classifier alignment phase for task \( t \), we randomly select \( N \) class prototypes (set to 32 in our implementation), denoted as \( \{\mu_{k}\}_{i=1}^{N}, k \in C^{1:t} \), within each training batch and generate pseudo-features \( \{\hat{\mu}_{k}\}_{i=1}^{N} \) via the PFR method. These pseudo-features, together with the training samples, are then used as inputs to the classifier, which is subsequently fine-tuned with the cross-entropy loss \( \mathcal{L}_{CE} \). 
{We provide a more detailed training pipeline in the supplementary material.}

\section{Experiments}

\begin{table*}[t!]
    \centering
    \renewcommand{\arraystretch}{0.7}
    \begin{tabular}{l r r *{11}{c}}
        \toprule\toprule
        \multirow{2}{*}{\textbf{Method}} & \multirow{2}{*}{\textbf{Params}} & \multirow{2}{*}{\textbf{Flops}} & \multicolumn{2}{c}{\textbf{ImageNet-R}} & \multicolumn{2}{c}{\textbf{ImageNet-A}} & \multicolumn{2}{c}{\textbf{CUB-200}} & \multicolumn{2}{c}{\textbf{Cifar-100}} \\
        \cmidrule(lr){4-5} \cmidrule(lr){6-7} \cmidrule(lr){8-9} \cmidrule(lr){10-11}
         &  &  & $\mathcal{A}^{10} \uparrow$ & $\bar{\mathcal{A}}^{10} \uparrow$ & $\mathcal{A}^{10} \uparrow$ & $\bar{\mathcal{A}}^{10} \uparrow$ & $\mathcal{A}^{10} \uparrow$ & $\bar{\mathcal{A}}^{10} \uparrow$ & $\mathcal{A}^{10} \uparrow$ & $\bar{\mathcal{A}}^{10} \uparrow$ \\
        \midrule\midrule
        FT              & 86M   & 17.58B  & 20.93 & 40.35 & 6.03 & 16.57 & 22.05 & 45.67 & 22.17 & 41.83 \\
        Adam-Ft
                        & 86M   & 33.72B  & 54.33 & 61.11 & 48.52 & 59.79 & 86.09 & 90.97 & 81.29 & 87.15 \\
        SLCA 
                        & 86M   & 17.58B  & 77.42 & 82.17 & \underline{60.63} & \underline{70.04} & 84.71 & 90.94 & \textbf{91.26} & \underline{94.09} \\
        \midrule
        Adam-Prompt-shallow
                        & 0.04M & 36.28B & 65.79 & 72.97 & 29.29 & 39.14 & 85.28 & 90.89 & 85.04 & 89.49 \\
        Adam-Prompt-deep
                        & 0.28M & 36.28B & 72.30 & 78.75 & 53.46 & 64.75 & \underline{86.6} & \underline{91.42} & 83.43 & 89.47 \\
        L2P
                        & 0.04M & 35.85B & 72.34 & 77.36 & 44.04 & 51.24 & 79.62 & 85.76 & 82.84 & 87.31 \\
        DualPrompt
                        & 0.13M & 33.72B & 69.10 & 74.28 & 53.19 & 61.47 & 81.10 & 88.23 & 83.44 & 88.76 \\
        CODA-Prompt
                        & 0.38M & 33.72B & 73.30 & 78.47 & 52.08 & 63.92 & 77.23 & 81.90 & 87.13 & 91.75 \\
        ConvPrompt
                        & 0.17M & 17.98B & \underline{77.86} & 81.55 & --- & --- & 82.44 & 85.59 & 88.10 & 92.39 \\
        CPrompt
                        & 0.25M & 23.62B & 77.15 & \underline{82.92} & 55.23 & 65.42 & 80.35 & 87.66 & 88.82 & 92.53 \\
        \midrule
        C-ADA
                        & 0.06M & 17.62B  & 73.76 & 79.57 & 54.10 & 65.43 & 76.13 & 85.74 & 88.25 & 91.85 \\
        LAE
                        & 0.19M & 35.24B  & 72.39 & 79.07 & 47.18 & 58.15 & 80.97 & 87.22 & 85.33 & 89.96 \\
        Adam-Adapter
                        & 1.19M & 36.47B  & 65.29 & 72.42 & 48.81 & 58.84 & 85.84 & 91.33 & 87.29 & 91.21 \\
        EASE
                        & 1.19M & 177.11B & 76.17 & 81.73 & 55.04 & 65.34 & 84.65 & 90.51 & 87.76 & 92.35 \\
        InfLoRA
                        & 0.19M & 21.79B  & 76.78 & 81.42 & 52.16 & 61.27 & 80.78 & 89.21 & 88.31 & 92.57 \\
        \midrule
        DIA (Ours)  & 0.17M & 17.91B & \textbf{79.03} & \textbf{85.61} & \textbf{61.69} & \textbf{71.58} & \textbf{86.73} & \textbf{93.21} & \underline{90.80} & \textbf{94.29} \\
        \bottomrule\bottomrule
    \end{tabular}%
    \caption{Experimental results on four CIL benchmarks with 10 incremental tasks. The best results are marked in \textbf{bold}, and the second-best are \underline{underlined}. We report the accuracy of compared methods with their source code. 
    \textbf{Params} indicates the average trainable parameters in each incremental task. \textbf{Flops} represents the number of floating-point operations required to perform inference on a single image.}
    \label{Table: Overview}
\end{table*}

In this section, we first provide implementation details, then present the experimental results with analyses, and finally show ablation studies and visualization results. 

\subsection{Implementation Details}
\label{Exp: implementation}
\subdivision{Dataset:}
We conduct experiments on Cifar-100~\cite{krizhevsky2009learning}, CUB-200~\cite{wah_branson_welinder_perona_belongie_2011}, ImageNet-R~\cite{ManyFacesRobustness_Hendrycks}, and ImageNet-A~\cite{Naturaladversarialexamples_Hendrycks}. 
These datasets contain typical CIL benchmarks and more challenging datasets with a significant domain gap with ImageNet (\textit{i.e.,} the pre-trained dataset). 
There are 100 classes in Cifar-100 and 200 classes in CUB-200, ImageNet-R, and ImageNet-A.
For all datasets, we follow the class orders in~\cite{2024CVPRExpandable_Zhou}. 

\subdivision{Comparison methods:} We compare our method with benchmark PTM-based CIL methods in \tab~\ref{Table: Overview} and \tab~\ref{Table: LongSeq}. These methods are categorized into three groups: 
(1) \textbf{Prompt-based methods:} L2P~\cite{2022CVPRLearning_Wang}, DualPrompt~\cite{2022ECCVDualPrompt_Wang}, CODA-Prompt~\cite{2023CVPRCODAprompt_Smitha}, CPrompt~\cite{2024CVPRConsistent_Gao}, ConvPrompt~\cite{2024CVPRConvolutional_Roy}, and Adam-Prompt~\cite{zhou2024revisiting} 
(2) \textbf{Adapter-based methods:} Adam-Adapter~\cite{zhou2024revisiting}, EASE~\cite{2024CVPRExpandable_Zhou}, LAE~\cite{2023ICCVunified_Gao}, InfLoRA~\cite{InfLoRAInterferencefreeLowrank_liang}, and C-ADA~\cite{PromptLearningContinual_gao}.
(3) \textbf{Finetuning-based methods:} SLCA~\cite{2023ICCVSLCA_Zhang} and Adam-Ft~\cite{zhou2024revisiting}.

\subdivision{Training details:} We adopt ViT-B/16~\cite{imageworth16x16_Dosovitskiy} as the pre-trained model, which is pre-trained on ImageNet-21K~\cite{ImageNetlargescale_Russakovskya}. The initial learning rate is set to 0.015 and decays with cosine annealing. We train each task for 20 epochs with a batch size of 32 using Nvidia 3090 GPUs with 24GB of RAM. The down projection of adapters is set to 8 in our main experiments. The hyperparameter \( \beta \) and \( \lambda \) are set to 0.7 and 0.1, respectively. 

\subdivision{Evaluation metric:} Following previous papers~\cite{2023ICCVunified_Gao,2024CVPRExpandable_Zhou}, we denote the model's accuracy after the \( t \)-th incremental task as \( A^t \) and use \( \bar{A}^{t}=\frac{1}{t}\sum_{i=1}^{t}{A^{i}} \) to represent the average accuracy over t incremental tasks. For our evaluation, we specifically focus on two key measurements: \( A^T \) (the accuracy after the final ($T$-th) incremental task) and \( \bar{A}^{T} \) (the average accuracy across all $T$ incremental tasks). 

\subsection{Comparison with SOTA Methods}
In this section, we evaluate our proposed DIA method on the ImageNet-R, ImageNet-A, CUB-200, and Cifar-100 datasets, equally dividing the classes into 10 incremental tasks. 
As shown in \tab~\ref{Table: Overview}, our method achieves SOTA average accuracies of 85.61\%, 71.58\%, 93.21\%, and 94.29\% on four benchmark datasets, respectively.

\subdivision{Prompt-based Methods:} 
In comparison to prompt-based approaches, the proposed DIA demonstrates comprehensive improvements. 
When compared with the current SOTA method CPrompt~\cite{2024CVPRConsistent_Gao}, our DIA exhibits accuracy enhancements ranging from 1.76\% to 4.99\% across all datasets. 
Moreover, the DIA requires fewer training parameters (merely 0.17M) and reduces the floating-point operations (FLOPS) per image by 24.17\%. 
ConvPrompt~\cite{2024CVPRConvolutional_Roy} improves inference efficiency by introducing large language models. Our approach, however, offers advantages in both computational efficiency and accuracy across all datasets, notably achieving 4.06\% higher average accuracy on ImageNet-R.

\subdivision{Adapter-based Methods:} 
In comparison to adapter-based methods, our DIA still performs exceptionally well. 
Utilizing only 14.28\% of the trainable parameters needed per incremental task, DIA outperforms EASE~\cite{2024CVPRExpandable_Zhou} and Adam~\cite{zhou2024revisiting} across all benchmark datasets.
Moreover, DIA only needs 17.91B Flops to infer an image, reducing inference consumption by 90\% compared to EASE~\cite{2024CVPRExpandable_Zhou} while improving accuracy. 
Specifically, DIA surpasses EASE~\cite{2024CVPRExpandable_Zhou} by 3.88\%, 6.24\%, 1.62\%, and 1.94\% on ImageNet-R, ImageNet-A, CUB-200, and Cifar-100, respectively. 
Our notable improvements on ImageNet-R and ImageNet-A underscore that incorporating compositionality and leveraging knowledge from old tasks can significantly enhance the learning of new classes that have a domain gap with the pre-trained data. 

\subdivision{Ft-based Methods:} 
Compared to finetuning-based methods, our DIA not only achieves SOTA average accuracy across four datasets but also saves 99.80\% of the training parameters required per task compared to SLCA and Adam-Ft. This demonstrates that our proposed DIA strikes an excellent balance between computational efficiency and performance.

\subsection{Ablation Study}

\begin{table}[t!]
    \centering
    \renewcommand{\arraystretch}{0.3}
    \begin{tabular}{l *{7}{c}}
        \toprule\toprule
        \multirow{2}{*}{\textbf{Method}} & \multicolumn{2}{c}{\textbf{ImageNet-R}} & \multicolumn{2}{c}{\textbf{Cifar-100}} \\
        \cmidrule(lr){2-3} \cmidrule(lr){4-5}
         &  $\mathcal{A}^{20} \uparrow$ & $\bar{\mathcal{A}}^{20} \uparrow$ & $\mathcal{A}^{20} \uparrow$ & $\bar{\mathcal{A}}^{20} \uparrow$ \\
        \midrule\midrule
        Adam-Ft
                         & 52.36 & 61.72  & 81.27 & 87.67 \\
        SLCA 
                         & 74.63 & 79.92 & \textbf{90.08} & \textbf{93.85} \\
        \midrule
        Adam-Prompt-shallow
                        & 59.90 & 68.02 & 84.57 & 90.43 \\
        Adam-Prompt-deep
                        & 70.13 & 76.91 & 82.17 & 88.46 \\
        L2P
                        & 69.64 & 75.28 & 79.93 & 85.94 \\
        DualPrompt
                        & 66.61 & 72.45 & 81.15 & 87.87 \\
        CODA-Prompt
                        & 69.96 & 75.34 & 81.96 & 89.11 \\
        ConvPrompt
                        & 74.3 & 79.66  & 87.25 & 91.46 \\
        CPrompt
                        & \underline{74.79} & \underline{81.46} & 84.57 & 90.51 \\
        \midrule
        LAE
                        & 69.86 & 77.38  & 83.69 & 87.86 \\
        Adam-Adapter
                        & 57.42 & 64.75  & 85.15 & 90.65 \\
        EASE
                        & 65.23 & 77.45 & 85.80 & 91.51 \\
        InfLoRA
                        & 71.01 & 77.28 & 85.80 & 91.51 \\
        \midrule
        DIA (Ours)  & \textbf{76.32} & \textbf{83.51} & \underline{88.74} & \underline{93.41} \\
        \bottomrule\bottomrule
    \end{tabular}%
    \caption{Experimental results with 20 incremental tasks. The best results are marked in \textbf{bold}, and the second-best are \underline{underlined}.}
    \label{Table: LongSeq}
\end{table}

\begin{table}[t!]
    \centering
    \renewcommand{\arraystretch}{0.3}
    \begin{tabular}{*{3}{l} *{4}{c}}
            \toprule\toprule
            \multirow{2}{*}{\textbf{DIA}} & \multirow{2}{*}{\textbf{PDL}} & \multirow{2}{*}{\textbf{PFR}} & \multicolumn{2}{c}{\textbf{ImageNet-R}} & \multicolumn{2}{c}{\textbf{Cifar-100}} \\
            \cmidrule(lr){4-5} \cmidrule(lr){6-7}
             &  &  & $\mathcal{A}^{10} \uparrow$ & $\bar{\mathcal{A}}^{10} \uparrow$ & $\mathcal{A}^{10} \uparrow$ & $\bar{\mathcal{A}}^{10} \uparrow$ \\
        \midrule\midrule
                   &            &            & 20.93 & 40.35 & 22.17 & 41.83 \\
        \checkmark &            &            & 77.13 & 83.87 & 88.37 & 92.31 \\
        \checkmark & \checkmark &            & 78.18 & 84.55 & 89.32 & 93.48 \\
        \checkmark &            & \checkmark & 78.22 & 84.25 & 89.81 & 93.78 \\
        \midrule
        \checkmark & \checkmark & \checkmark & 79.03 & 85.61 & 90.80 & 94.29 \\
        \bottomrule\bottomrule
    \end{tabular}%
    \caption{Ablation Study on Different Components of the DIA Framework.}
    \label{Table: Ablation}
\end{table}

\subdivision{Long Sequence Incremental Analysis:}
We evaluate the performance of each method under the long sequence setting where datasets are equally divided into 20 tasks with 10 classes/tasks in ImageNet-R and 5 classes/tasks in Cifar-100. As shown in \tab~\ref{Table: LongSeq}, our method still maintains excellent performance in terms of \( A^{20} \) and \( \bar{A}^{20} \). 
Our DIA achieves SOTA average accuracies of 83.51\% on the ImageNet-R dataset, outperforming the second by 2.05\%. Our DIA also achieve comparable results with SLCA on Cifar-100, using only 0.17M/86M parameters. 

\subdivision{Component Analysis:}
Our DIA comprises three main components: 1) Task-Specific Adapter Integration (TSAI), 2) Patch-Level Distillation Loss (PDL), and 3) Patch-Level Feature Reconstruction (PFR). To assess the impact of each component, we perform ablation studies on Cifar-100 and ImageNet-R. 

The first row in \tab~\ref{Table: Ablation} reports the average and final accuracy of the pre-trained image encoder with a learnable classifier. Adding the TSAI module achieves performance comparable to SOTA on both datasets, showing that patch-level integration reduces task interference and improves new task learning. The third and fourth rows further demonstrate the importance of patch-level model alignment in NECIL, with nearly 1\% gains from PDL or PFR integration. Finally, the fifth row confirms that combining compositionality and model alignment yields SOTA results, validating our framework’s effectiveness.

\begin{table}[t!]
    \centering
    \renewcommand{\arraystretch}{0.3}
    \begin{tabular}{p{0.15\textwidth} l*{3} {c}}
        \toprule\toprule
        \multirow{2}{*}{\textbf{Method}} & \multicolumn{2}{c}{\textbf{ImageNet-R}} & \multicolumn{2}{c}{\textbf{Cifar-100}} \\
        \cmidrule(lr){2-3} \cmidrule(lr){4-5}
         & $\mathcal{A}^{10} \uparrow$ & $\bar{\mathcal{A}}^{10} \uparrow$ & $\mathcal{A}^{10} \uparrow$ & $\bar{\mathcal{A}}^{10} \uparrow$ \\
        \midrule\midrule
        DIA-Gau     & 76.07 & 83.72 & 89.04 & 93.51 \\
        DIA-SLCA    & 78.37 & 84.48 & 89.47 & 93.79 \\
        \midrule
        DIA-PFR     & 79.03 & 85.61 & 90.80 & 94.29 \\
        \bottomrule\bottomrule
    \end{tabular}
    \caption{Performance comparison with different feature reconstruction (FR) methods.}
    \label{Table: Rehearsal}
\end{table}

\definecolor{skyblue}{HTML}{87CEEB}
\definecolor{darkblue}{HTML}{000080}

\begin{figure*}[t!]
    \centering
    \includegraphics[width=1.0\linewidth]{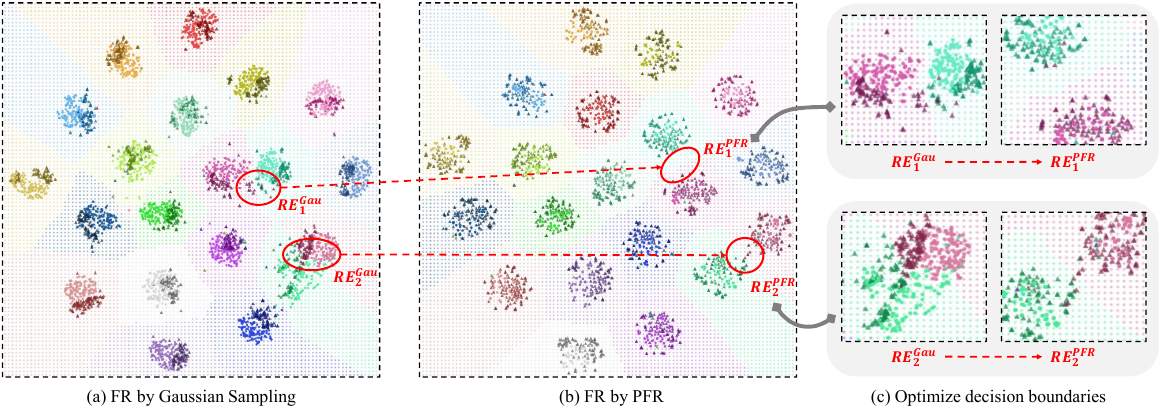}
    \caption{\textbf{Visualization of different feature reconstruction methods.} Dark-colored triangles represent actual features, while light-colored circles denote pseudo features generated by feature reconstruction (FR) methods. The decision boundaries formed by these pseudo features are also visualized through the background color. \( RE^{Gau} \) and \( RE^{PFR} \) indicate regions with inaccurate decision boundaries. The features reconstructed by PFR closely align with the actual features of old classes, resulting in more accurate decision boundaries.}
    \label{fig: visual}
\end{figure*}

\begin{table}[t]
    \centering
    \renewcommand{\arraystretch}{0.3}
    \begin{tabular}{l *{4}{c}}
            \toprule\toprule
            \multirow{2}{*}{\textbf{Distillation} \textbf{Loss}} & \multicolumn{2}{c}{\textbf{ImageNet-R}} & \multicolumn{2}{c}{\textbf{Cifar-100}} \\
            \cmidrule(lr){2-3} \cmidrule(lr){4-5}
             & $\mathcal{A}^{10} \uparrow$ & $\bar{\mathcal{A}}^{10} \uparrow$ & $\mathcal{A}^{10} \uparrow$ & $\bar{\mathcal{A}}^{10} \uparrow$ \\
        \midrule\midrule
        \(\mathcal{L}_{pdl} \) w/ \( \alpha_{\text{eu}} \)  & 76.38 & 81.73 & 87.42 & 92.91 \\
        \(\mathcal{L}_{pdl} \) w/ \( \alpha_{\text{cos}} \)    & 78.28 & 85.16 & 89.67 & 93.88 \\
        \(\mathcal{L}_{pdl} \) w/ \( \alpha_{\angle} \) (ours) & 79.03 & 85.61 & 90.80 & 94.29 \\
        \midrule
        \(\mathcal{L}_{fd} \)               & 76.67 & 83.82 & 88.31 & 92.97 \\
        \(\mathcal{L}_{pdl} \) w/ cls       & 77.01 & 83.87 & 88.87 & 93.02 \\
        \(\mathcal{L}_{pdl} \) w/o cls (ours)      & 79.03 & 85.61 & 90.80 & 94.29 \\
        \bottomrule\bottomrule
    \end{tabular}%
    \caption{Ablation Study on Similarity Metrics in \( \mathcal{L}_{pdl} \) and other distillation methods.}
    \label{Table: Metric}
\end{table}

\begin{figure}[t!]
    \centering
    \includegraphics[width=1.0\linewidth]{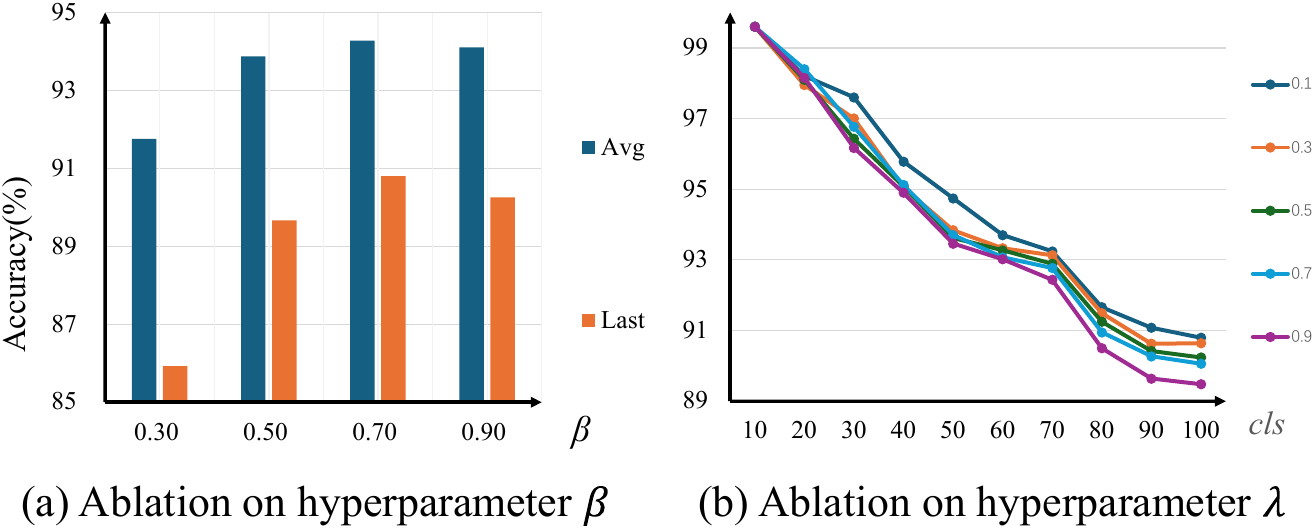}
    \caption{Ablation experiments on hyperparameters \( \beta \) and \( \lambda \) conducted on the CIFAR-100 dataset.}
    \label{fig: ab.hp}
\end{figure}

\subdivision{Feature Reconstruction Analysis:}
We investigate the impact of different pseudo-feature reconstruction methods on classifier alignment. 
As shown in \tab~\ref{Table: Rehearsal}, constructing old class features using a Gaussian distribution results in accuracies of 83.72\% and 93.51\% on ImageNet-R and CIFAR-100, respectively. 
In contrast, our proposed PFR method leverages rich semantics from patch tokens to achieve average accuracies of 85.61\% and 94.29\%, outperforming other approaches. This demonstrates the effectiveness of PFR in aligning the classifier with old-class features, leading to more accurate decision boundaries.

{We also visualize the pseudo-feature distributions generated by Gaussian sampling and compare them to the actual old class features on new tasks, as shown in \fig\ref{fig: visual}. 
After learning the last incremental task, we visualize the feature distribution of the first 20 classes. \fig\ref{fig: visual} (a) shows pseudo features generated using a Gaussian distribution, forming two distinct clusters with actual features, highlighting a noticeable disparity. In contrast, \fig\ref{fig: visual} (b) displays pseudo features generated by PFR, which closely align with actual features, forming a single cohesive cluster. This further results in a more clearly distinguishable decision boundary, as depicted in \fig\ref{fig: visual} (b) and (c), thereby confirming the effectiveness of our proposed feature reconstruction method.}

\subdivision{PDL Ablation:}
We explore different similarity metrics to evaluate patch token contributions to new task learning and their impacts on PDL. Specifically, we investigate Euclidean distance (\( \alpha_{\text{eu}} \))~\cite{2024AAAIFineGrained_zhai}, cosine similarity (\( \alpha_{\text{cos}} \)), and angular similarity (\( \alpha_{\angle} \)) as shown in \tab~\ref{Table: Metric}. The experimental results indicate angular similarity outperforms others, achieving SOTA results on both datasets. In contrast, Euclidean distance performs the worst, with a significant accuracy drop. We attribute this poor performance to the numerical instability of the Euclidean distance, which is more sensitive to variations in the numerical range and more prone to outliers.

Additionally, ablation experiments on traditional distillation loss are presented. In \tab~\ref{Table: Metric}, \( \mathcal{L}_{fd} \)~\cite{2022SSRE,2021CVPRPrototype_Zhu,2023ACMMPolo_wang} refers to applying \( L1 \) constraints on all image tokens, while \( \mathcal{L}_{pdl} \) w/ cls~\cite{2024AAAIFineGrained_zhai} applies the PDL constraint to both the class and patch tokens. The results show that our proposed PDL offers a better trade-off between stability (regularization of feature shift) and plasticity (ability to learn new tasks).

\subdivision{Hyperparameter Ablation}
We conduct ablation studies for the hyperparameters \( \beta \) and \( \lambda \) in Fig.\ref{fig: ab.hp}. 
As depicted in the figure, larger \( \lambda \) which results in greater regularization hinders the learning process. Small \( \beta \) may lead to inaccurate classifier alignment. Therefore, we suggest the optimal hyperparameter combination of \( \beta = 0.7 \) and \( \lambda = 0.1 \).

\section{Conclusion}
In this paper, we propose a novel framework called Dynamic Integration of task-specific Adapters (DIA), which consists of Task-Specific Adapter Integration (TSAI) and Patch-Level Model Alignment.
First, the TSAI module enhances compositionality through a patch-level adapter integration mechanism, minimizing task interference while preserving old task knowledge. 
Afterward, the Patch-level Model Alignment maintains feature consistency and accurate decision boundaries via a patch-level distillation loss and a patch-level feature reconstruction method, respectively. 
1) Our PDL preserves feature-level consistency between successive models by implementing a distillation loss based on the contribution of patch tokens to new class learning. 
2) Our PFR facilitates accurate classifier alignment by reconstructing features from previous tasks that adapt to new task knowledge. 

\clearpage

\section{Acknowledgment}
{
This work was supported by the National Key Research and Development Project of China (Grant 2020AAA0105600), the National Natural Science Foundation of China (Grant No. U21B2048), Shenzhen Key Technical Projects (Grant CJGJZD2022051714160501) and Natural Science Foundation of Shaanxi Province 2024JC-YBQN-0637.
}

{
    \small
    \bibliographystyle{ieeenat_fullname}
    \bibliography{main}
}

\end{document}


\clearpage
\maketitlesupplementary

\section{Methodology Supplementary}
\subsection{Analysis of knowledge of retention and reproduction:}
As mentioned in the main manuscript, for an input token \( \mathbf{p} \in \mathbb{R}^{m} \), consider an adapter without an activation function, with weights \( W = \mathbf{W}_{\text{down}}\mathbf{W}_{\text{up}} \in \mathbb{R}^{m \times n},\mathbf{W}_{\text{down}} \in \mathbb{R}^{m \times r}, \mathbf{W}_{\text{up}} \in \mathbb{R}^{r \times n} \), and a matrix rank of \( r \). The weight matrix can be decomposed using SVD as follows: 
\begin{align}
    \mathbf{W} &= \mathbf{U} \text{diag}(\sigma) \mathbf{V}, \\
    \mathbf{U}^\top = \left[ {\mathbf{u}_i}^\top \right]_{i=1}^r & \in \mathbb{R}^{r \times m},
    \mathbf{V} = {\left[ {\mathbf{v}_i}^\top \right]_{i=1}^r} \in \mathbb{R}^{r \times n}.
\label{eq: SVD}
\end{align}
And the output \( \mathbf{o} \) can be formulated as:
\begin{align}
     o &= \mathcal{A}(\mathbf{p}) = \mathbf{W}^\top \mathbf{p} = \sum {\rho}^{'}_{i} \mathbf{v}_i = \mathbf{V}^\top g(\mathbf{p}), \\
     g(\mathbf{p}) &= {\text{diag}(\sigma)}^\top \mathbf{U}^\top \mathbf{p}= {[\rho^{'}_{i}]_{i=1}^r} \in \mathbb{R}^r.
     \label{eq: linear}
\end{align}

After introducing non-linear activation function \( \text{ReLU} \), for each input \( \mathbf{p} \in \mathbb{R}^m \), the output of the adapter can be reformulated as:
\begin{align}
\mathcal{A}(\mathbf{p}) = \mathbf{W}_{\text{up}}^T \text{ReLU}(\mathbf{W}_{\text{down}}^T\mathbf{p}), \ \end{align}
We perform SVD decomposition on \( \mathbf{W}_{\text{up}} \) as the above Eq.~\ref{eq: SVD}:
\begin{align}
    & \mathbf{\mathbf{W}_{\text{up}}} = \mathbf{U}_{\text{up}} \text{diag}(\sigma_{\text{up}}) \mathbf{V}_{\text{up}}, \\
    & \mathbf{W}_{\text{up}} = \sum \mathbf{u}_{(i, {\text{up}})} \sigma_{(i,{\text{up}})} \mathbf{v}_{(i,{\text{up}})}^\top.
\end{align}
The output \( \mathbf{o} \) can be reformulated as:
\begin{align}
 o  &= \mathcal{A}(\mathbf{p})= \mathbf{W}_{\text{up}}^T \text{ReLU}(\mathbf{W}_{\text{down}}^T\mathbf{p}) \nonumber \\
    &= \sum {(\mathbf{u}_{(i, {\text{up}})} \sigma_{(i,{\text{up}})} \mathbf{v}_{(i,{\text{up}})}^\top)}^\top \text{ReLU}(\mathbf{W}_{\text{down}}^T\mathbf{p}) \nonumber  \\
   &= \sum \mathbf{v}_{(i,{\text{up}})} \left( \sigma_{(i,{\text{up}})} \mathbf{u}_{(i, \text{up})}^\top \text{ReLU}(\mathbf{W}_{\text{down}}^T\mathbf{p}) \right) \nonumber \\
   &= \sum \rho^{'}_{i} \mathbf{v}_{(i,{\text{up}})} = \mathbf{V}_{\text{up}}^\top g(\mathbf{p}), \\
 g(\mathbf{p}) &= {\text{diag}(\sigma_{(i,{\text{up}})})}^\top \mathbf{U}_{(i,{\text{up}})}^\top \text{ReLU}(\mathbf{W}_{\text{down}}^T\mathbf{p}) \in \mathbb{R}^r.
 \label{eq: non-linear}
\end{align}
From Eq.~\ref{eq: non-linear}, we observe that after introducing the nonlinear activation, we obtain an expression similar to Eq.~\ref{eq: linear}, with the function \( g \) replaced by a nonlinear function.

\begin{algorithm}[t!]
\caption{Training Pipeline for Task \(t\)}
\textbf{Input}: Training dataset \(D^t\); TSAI module \( f(\cdot;\theta_b) \); Cosine classifier \( \phi(\cdot; \mathbf{W}_{cls}) \); \\
\textbf{Parameter}: \( \theta_b = \{\theta_{ptm}, \theta^o, \theta^n \} \), \(\mathbf{W}_{cls}=\{ \mathbf{W}_{cls}^o,\mathbf{W}_{cls}^n \} \); \\
\textbf{Initialization}: Initialize \( \theta^n \) and \( \mathbf{W}_{cls}^n \). Freeze parameters \( \theta_{ptm} \) and \( \theta^o \);
\begin{algorithmic}[1] 
\STATE\# New Task Learning.
\WHILE{not converged}
    \FOR{\( \{I_i^t, y_i^t\} \in D^t\)}
        \STATE Compute logits \( \xi_{y_i^t} =\phi \left( f(I_i^t; \theta_{b}); \mathbf{W}_{cls}^n \right) \);
        \STATE Compute CE loss \( \mathcal{L}_{CE}(\xi_{y_i^t}; {s}, \mathfrak{m}) \);
        \STATE Compute patch-level distillation loss \(\mathcal{L}_{pld}\);
        \STATE Backward with objective \( \mathcal{L}_{obj} = \mathcal{L}_{CE} + \lambda \mathcal{L}_{PDL} \);
    \ENDFOR
\ENDWHILE
\STATE For each class \( c_k^t \in C^t \) in task \( t \), we compute its class prototype \( \bm{\mu}_k^t\) and store it in memory;
\STATE\# Classifier alignment.
\WHILE{not converged}
    \FOR{Training batch within \( D^t\)}
        \STATE Sample \textit{N} class prototypes \( \{\mu_{c_i}\}_{i=1}^{N}, c_i \in C^{1:t} \);
        \STATE Construct class feature \( \hat{\mu}_{i, c} \) using PFR method with the training batch;
        \STATE Fine-tune the classifier \( \phi(\cdot; \mathbf{W}_{cls}^o,\mathbf{W}_{cls}^n) \);
    \ENDFOR
\ENDWHILE
\end{algorithmic}
\label{alg: pipline}
\end{algorithm}

\subsection{Training Pipeline}
The training pipeline for the proposed DIA method follows previous approaches~\cite{2023ICCVSLCA_Zhang,2023ACMMPolo_wang,2021CVPRPrototype_Zhu} and is composed of two stages: new task learning and classifier alignment for each task \( t \), as illustrated in Algorithm.\ref{alg: pipline}.

For incremental task \( t \), the TSAI module is parameterized by \( \theta_b = \{\theta_{ptm}, \theta^o, \theta^n \} \), where \( \theta_{ptm} \) denotes the parameters of the pre-trained model (PTM), \( \theta^o \) refers to the parameters of the adapters and signature vectors for old tasks, and \( \theta^n \) corresponds to the parameters for the new task \( t \). The cosine classifier \(\phi(\cdot; \mathbf{W}_{cls})\) comprises two parts: the old class classifier \(\phi^o(\cdot; \mathbf{W}_{cls}^o)\) and the new class classifier \(\phi^n(\cdot; \mathbf{W}_{cls}^n)\).

\begin{table}[t!]
    \centering
    \begin{tabular}{l *{5}{c}}
        \toprule\toprule
        \multirow{2}{*}{\textbf{Method}}  & \multirow{2}{*}{\textbf{Backbone}} & \multicolumn{2}{c}{\textbf{ImageNet-R}} & \multicolumn{2}{c}{\textbf{Cifar-100}} \\
        \cmidrule(lr){3-4} \cmidrule(lr){5-6}
         & & $\mathcal{A}^{10} \uparrow$ & $\bar{\mathcal{A}^{10}} \uparrow$ & $\mathcal{A}^{10} \uparrow$ & $\bar{\mathcal{A}^{10}} \uparrow$ \\
        \midrule\midrule
        RAPF & CLIP & \textbf{80.28} & \underline{85.58} & 79.04 & \underline{86.19} \\
        P-Fusion & CLIP & 79.10 & - & \underline{85.50} & - \\
        DIA-r16 & ViT-B16 & \underline{79.82} & \textbf{86.04} & \textbf{90.38} & \textbf{94.36 }\\
        \bottomrule\bottomrule
    \end{tabular}%
    \caption{Comparison with CLIP-based CIL methods with their reported accuracy. The best results are marked in \textbf{bold}, and the second are marked in \underline{underline}.}
    \label{Table: clip}
\end{table}

\begin{table*}[t!]
    \centering
    \begin{tabular}{l *{10}{c}}
        \toprule\toprule
        \multirow{2}{*}{\textbf{Method}} & \multirow{2}{*}{\textbf{Params}} & \multirow{2}{*}{\textbf{Flops}} & \multicolumn{2}{c}{\textbf{ImageNet-R}} & \multicolumn{2}{c}{\textbf{ImageNet-A}} & \multicolumn{2}{c}{\textbf{CUB-200}} & \multicolumn{2}{c}{\textbf{Cifar-100}} \\
        \cmidrule(lr){4-5} \cmidrule(lr){6-7} \cmidrule(lr){8-9} \cmidrule(lr){10-11}
         &  &  & $\mathcal{A}^{10} \uparrow$ & $\bar{\mathcal{A}^{10}} \uparrow$ & $\mathcal{A}^{10} \uparrow$ & $\bar{\mathcal{A}^{10}} \uparrow$ & $\mathcal{A}^{10} \uparrow$ & $\bar{\mathcal{A}^{10}} \uparrow$ & $\mathcal{A}^{10} \uparrow$ & $\bar{\mathcal{A}^{10}} \uparrow$ \\
        \midrule\midrule
        DIA-MLP & 0.17M & 17.91B & 79.03 & 85.61 & 61.69 & 71.58 & 86.73 & 93.21 & 90.8 & 94.28 \\
        DIA-MHSA & 0.51M & 18.56B & 78.50 & 85.05 & 57.27 & 69.12 & 83.29 & 90.56 & 90.33 & 94.17 \\
        DIA-MIX & 0.69M & 18.9B & 79.69 & 85.41 & 58.21 & 69.93 & 85.68 & 92.38 & 90.85 & 94.30 \\
        \bottomrule\bottomrule
    \end{tabular}%
    \caption{Ablation experiments on the adapter structure with 10 incremental tasks.}
    \label{Table: Structure}
\end{table*}

\begin{table*}[t!]
    \centering
    \begin{tabular}{l *{10}{c}}
        \toprule\toprule
        \multirow{2}{*}{\textbf{Method}} & \multirow{2}{*}{\textbf{Params}} & \multirow{2}{*}{\textbf{Flops}} & \multicolumn{2}{c}{\textbf{ImageNet-R}} & \multicolumn{2}{c}{\textbf{ImageNet-A}} & \multicolumn{2}{c}{\textbf{CUB-200}} & \multicolumn{2}{c}{\textbf{Cifar-100}} \\
        \cmidrule(lr){4-5} \cmidrule(lr){6-7} \cmidrule(lr){8-9} \cmidrule(lr){10-11}
         &  &  & $\mathcal{A}^{10} \uparrow$ & $\bar{\mathcal{A}^{10}} \uparrow$ & $\mathcal{A}^{10} \uparrow$ & $\bar{\mathcal{A}^{10}} \uparrow$ & $\mathcal{A}^{10} \uparrow$ & $\bar{\mathcal{A}^{10}} \uparrow$ & $\mathcal{A}^{10} \uparrow$ & $\bar{\mathcal{A}^{10}} \uparrow$ \\
        \midrule\midrule
        DIA-r08 & 0.17M & 17.91B & 79.03 & 85.61 & 61.69 & 71.58 & 86.73 & 93.21 & 90.8 & 94.28 \\
        DIA-r16 & 0.31M & 18.18B & 79.82 & 86.04 & 57.74 & 69.84 & 86.41 & 92.39 & 90.38 & 94.36 \\
        DIA-r64 & 1.19M & 19.20B & 79.10 & 85.62 & 59.91 & 70.90 & 87.19 & 92.26 & 90.56 & 94.51 \\
        \bottomrule\bottomrule
    \end{tabular}%
    \caption{Ablation experiments on the adapter rank with 10 incremental tasks.}
    \label{Table: rank}
\end{table*}

\subdivision{New Task Learning:}
During new task learning, for each image \( \{I_i^t, y_i^t\} \in D^t\), where \( y_i^t \in c_k^t \in C^t \), we first extract features \( f(I_i^t; \theta_{b}) \) using TSAI module \( f(\cdot; \theta_{b}) \), then compute the output logits 
\begin{align}
\xi_{y_i^t} =\phi^n \left( f(I_i^t; \theta_{b}); \mathbf{W}_{cls}^n \right), 
\end{align}
with the cosine classifier of new task \( \phi^n(\cdot; \mathbf{W}_{cls}^n) \). We optimize the classification results using a variant of the CE loss \( \mathcal{L}_{CE} \) and ensure feature consistency with patch-level distillation loss (PDL) \(\mathcal{L}_{pdl}\).

Following methods~\cite{peng2022few,wang2018cosfacelargemargincosine}, we define \( \mathcal{L}_{CE} \) as:
\begin{align}
    \mathcal{L}_{CE}(\xi_{y_i^t}; {s}, \mathfrak{m}) =
    -\text{log}{
    \frac{e^{s( \xi_{y_i^t}-\mathfrak{m} )}}
    {e^{s( \xi_{y_i^t}-\mathfrak{m} )}+\sum_{c}^{C^{t}-{c_i^t}}{e^{s( \xi_{c} )}}}
    }.
\end{align}
Here, \( s \) is a scaling factor that adjusts the magnitude of cosine similarity, ensuring that the Softmax function produces a more discriminative probability distribution. \( \mathfrak{m} \) introduces an additional angular separation between classes. When \( s = 0 \) and \( \mathfrak{m} = 0 \), \( \mathcal{L}_{CE}(\cdot; 0, 0) \) reduces to the standard cross-entropy loss function. When \( s \neq 0 \) and \( \mathfrak{m} = 0 \), \( \mathcal{L}_{CE}(\cdot; s, 0)\) adopts a common format used in cosine classifiers, adjusting the scale of cosine similarity via \( s \).

The loss function during training is defined as:
\begin{align}
    \mathcal{L}_{obj} = \mathcal{L}_{CE} + \lambda \mathcal{L}_{pdl},
\end{align}
where \( \lambda \) is a hyperparameter that controls the strength of the regularization.

\subdivision{Classifier Alignment:}
To further refine the classification layer, we perform classifier alignment after the new task learning stage, following methods~\cite{2023ICCVSLCA_Zhang,2021CVPRPrototype_Zhu} (see Algorithm 1). Specifically, during the new task learning phase, only the classifiers corresponding to the current task's classes are trained alongside the TSAI module. Once new task learning is completed, we compute the class prototype \( \bm{\mu}_k^t = \frac{1}{N^t_k}\sum_{i}^{N^t_k}{f(I_i^t;\theta_{b})} \) for each class \( c_k^t \in C^t \) in the current task, where \( N^t_k \) denotes the number of images for class \( c_k^t \).

During the classifier alignment stage for task \( t \), we randomly select \( N \) (Set to 32 in our implementation) class prototypes \( \{\mu_{c_i}\}_{i=1}^{N}, c_i \in C^{1:t} \) in each training batch and generate pseudo-features \( \{\hat{\mu}_{c_i}\}_{i=1}^{N} \) using the PFR method. These pseudo-features, combined with training samples, are used as inputs to the classifier \(\phi(\cdot; \mathbf{W}^o_{cls}, \mathbf{W}^n_{cls})\), which is then fine-tuned using \( \mathcal{L}_{CE}(\cdot; 0, 0) \).

\section{Supplementary Experiments}
This section provides additional experiments, including comparisons with CLIP-MoE~\cite{yu2024boosting}, hyperparameter ablation studies, and model structure ablation studies.

\subsection{Comparative Experiments}
We conduct comparative experiments with the latest methods that use a CLIP backbone: RAPF~\cite{ClassIncrementalLearningCLIP_huang} and P-Fusion~\cite{PromptFusionDecouplingStability_chen}. As shown in Table~\ref{Table: clip}, despite the advantage of a more powerful CLIP backbone and access to additional semantic information, RAPF~\cite{ClassIncrementalLearningCLIP_huang} and P-Fusion~\cite{PromptFusionDecouplingStability_chen} are outperformed by our method on the CIFAR-100 dataset by a significant margin. Furthermore, our approach achieves comparable performance to these methods on the Imagenet-R dataset.

\subsection{Structure Ablation}
We explore the impact of inserting the TSAI module in parallel within both the MHSA and MLP structures of the transformer block. Specifically, we integrate TSAI in parallel with the three QKV projection layers of MHSA. As shown in Table~\ref{Table: Structure}, despite tripling the number of trainable parameters per task, DIA-MHSA still slightly underperforms compared to DIA-MLP. We attribute this to the multi-head attention mechanism's role in capturing input sequence dependencies, where its complex structure may be disrupted by the addition of adapters, thereby increasing optimization difficulty. The experimental results further validate the rationality of our current model structure.

\subsection{Adapter Rank Ablation}
We evaluate the model's accuracy across four datasets by varying the adapter down-projection dimensions to 8, 16, and 64 in Table~\ref{Table: rank}. The results show that increasing the number of parameters does not lead to significant accuracy improvements, with the r64 configuration yielding less than a one-point gain over r08. This demonstrates that the importance of model architecture and training strategies outweighs that of merely increasing the number of parameters.

\begin{table}[t!]
    \centering
    \begin{tabularx}{0.42\textwidth}{lcccc} 
        \toprule\toprule
        \multirow{2}{*}{\textbf{Ablation}} & 
        \multicolumn{2}{c}{\textbf{ImageNet-R}} & 
        \multicolumn{2}{c}{\textbf{Cifar100}} \\
        \cmidrule(lr){2-3} \cmidrule(l){4-5}
        & {$\mathcal{A}^{10}$} & {$\bar{\mathcal{A}}^{10}$} & 
        {$\mathcal{A}^{10}$} & {$\bar{\mathcal{A}}^{10}$} \\
        \midrule\midrule
        DIA w SDC & 75.13 & 83.63 & 89.13 & 93.46 \\
        DIA w LDC & 76.43 & 84.08 & 88.66 & 93.66 \\
        DIA w PFR & 79.03 & 85.61 & 90.80 & 94.29 \\
        \bottomrule\bottomrule
    \end{tabularx}
    \caption{Ablation experiments on the feature shift techniques with 10 incremental tasks.}
    \label{Table: FD}
\end{table}

\subsection{Feature Drift Discussion}
We further perform ablation studies to evaluate the impact of feature shift techniques on model alignment, as summarized in Table~\ref{Table: FD}
The results demonstrate that compared to SDC~\cite{SDC2020}, which calculates class prototype shifts using feature gaps, or LDC~\cite{LDC2024}, which uses MLP to learn mappings from old to new feature spaces, our proposed PFR method better captures the distribution of old class features in the new task's feature space, achieving superior accuracy on both ImageNet-R and CIFAR-100 datasets.

{
    \small
    \bibliographystyle{ieeenat_fullname}
    \bibliography{main}
}